# A Deep Dive into the Factors Influencing Financial Success: A Machine Learning Approach

Michael Zhou[1], Ramin Ramezani[2]
[1]Interlake High School, Bellevue, WA, michaelzhou2025@gmail.com
[2]University of California, Los Angeles, California, raminr@ucla.edu

# Abstract

This paper explores various socioeconomic factors that contribute to individual financial success using machine learning algorithms and approaches.

Financial success, a critical aspect of all individual's well-being, is a complex concept influenced by various factors. This study aims to understand the determinants of financial success. It examines the survey data from the National Longitudinal Survey of Youth 1997 by the Bureau of Labor Statistics (1), consisting of a sample of 8,984 individuals's longitudinal data over years. The dataset comprises income variables and a large set of socioeconomic variables of individuals.

An in-depth analysis shows the effectiveness of machine learning algorithms in financial success research, highlights the potential of leveraging longitudinal data to enhance prediction accuracy, and provides valuable insights into how various socioeconomic factors influence financial success.

The findings highlight the significant influence of highest education degree, occupation and gender as the top three determinants of individual income among socioeconomic factors examined. Yearly working hours, age and work tenure follow as three secondary influencing factors, and all other factors including parental household income, industry, parents' highest grade and others are identified as tertiary factors.

These insights allow researchers to better understand the complex nature of financial success, and are also crucial for fostering financial success among individuals and advancing broader societal well-being by providing insights for policymakers during decision-making process.

# Introduction

In the contemporary world, financial success and achieving financial success has gradually gained more and more weight in the minds of all individuals. As a result of the capitalist economy under which much of the world now operates on, the notion of financial success, which once was a personal ambition, has become a worldly phenomenon with many far-reaching implications. Recognizing and navigating the intricacies of financial success is one of the most paramount endeavors that researchers must take in order to grasp the force that shapes aspirations, decision-making, and the broader socio-economic fabric of society.

To better understand the complexities associated with financial success and income analysis, we will start off by defining some terms. "Financial success" in the context of this study encompasses the monetary gains of individuals by the same individuals in a one year time-frame.

"Income analysis" refers to the systematic evaluation of patterns between those who earn a higher income and those who earn a lower income. The research of these key concepts will help dissect the complex nature of financial success and provide nuanced insights into the factors shaping income dynamics.

## Objectives

In this big data era, advanced data science has been increasingly used in many industries, including economic research. There is a growing interest in utilizing machine learning methods in economic research, surpassing traditional statistical models. Papers (2-5) in Literature Review are some examples using machine learning methods in economic research.

One prominent effort in this field of study is to predict individual income based on a range of socioeconomic factors. These factors typically include education, employment, demographic information, socioeconomic status, and other relevant variables. Papers (6-13) in Literature Review provide illustrative examples in this research area.

In this study, we exploit the data from National Longitudinal Survey of Youth 1997 (hereinafter abbreviated as NLSY97) to conduct a multi-class classification task at predicting individual income levels with machine learning algorithms. The primary purpose is to identify dominant socioeconomic factors that influence an individual's income. Various socioeconomic factors, including demographic, educational, and occupational factors, are examined for their influence on individual income levels. This list is not exhaustive but serves as a starting point for further exploration. Each factor targets a different aspect of an individual's life as it relates to financial success. When combined, these factors provide a more nuanced influence than any single factor alone. Through the analysis of these factors, this study provides a holistic view of their influence on individual economic well-being and financial success, both independently and collectively. Furthermore, we want to explore the longitudinal data in the survey data to enhance prediction accuracy. In contrast to existing research on individual income prediction, which primarily focuses on exploring population characteristics in the dataset, this study also explores trajectories of individuals over time in the longitudinal data to improve the accuracy of prediction tasks, yielding satisfactory results.

## Literature Review

Here are some notable instances of previous economic research that effectively employed machine learning methods, with successful outcomes.

Yeh, C., Perez, A., Driscoll, A. et al (2) used a satellite-based deep learning approach to predict economic well-being in Africa with satisfactory results.

Sheetal, A., Chaudhury, S.H. and Savani, K (3) demonstrated that machine learning methods can be used to predict people's attitudes toward income inequality and their experiments revealed that a greater emphasis on hard work was associated with a greater justification of income inequality.

R. Kannan, K. W. Shing, K. Ramakrishnan, H. B. Ong and A. Alamsyah (4) showed that the machine learning approach can provide a robust model to predict households' financial vigilances.

Aiken, E., Bellue, S., Karlan, D. et al. (5) used traditional survey data to train machine-learning algorithms to recognize patterns of poverty in mobile phone data and reduced errors of exclusion by 4–21%, highlighting potential for targeting humanitarian assistance.

Before delving into the specifics of our study on individual income prediction with machine learning methods, it is essential to review key findings and gaps in the existing literature. Previous research has explored various socioeconomic factors related to financial success using machine learning methodologies. Here are some representative studies in this specific area.

Kamdjou, Herve D. Teguim (6) estimated the income returns to education using a machine learning approach and concluded that each additional year of education, on average, yields a private rate of returns of 10.4%.

Gomez-Cravioto, D.A., Diaz-Ramos, R.E., Hernandez-Gress, N. et al (7) employed supervised machine learning predictive analytics to predict alumni income. Their study centered on survey data collected in 2018 by Tecnologico de Monterrey and yielded statistically significant results ($p < 0.05$), with particular accuracy in predicting the alum's first income after graduation.

A. Lazar (8) employed principal component analysis and support vector machine methods to generate and evaluate income prediction data based on the Current Population Survey provided by the U.S. Census Bureau and achieved accuracy values as high as 84% for binary classification tasks.

Matz SC, Menges JI, Stillwell DJ, Schwartz HA (9) applied an established machine learning method to predict individual income from the digital footprints people leave on Facebook with an accuracy up to 0.49.

P. Khongchai and P. Songmuang (10) proposed a salary prediction system based on decision tree technique with seven features to increase students' motivation in studying. It yielded overall 41.39% accuracy and had positive satisfaction with 50 student samples.

Islam Rana, Cp Md et al. (11) investigated income prediction performance of eleven machine learning models on UCI Adult Dataset and showed that demographic characteristics help some models predict income levels better.

Zaid, Mohamed and RajendranT (12) developed a decision tree algorithm with a high mean accuracy of 84.3790 on predicting income class for "adult income dataset" acquired with open source Kaggle platform.

Chakrabarty, Navoneel and Biswas, Sanket (13) developed a Gradient Boosting Classifier model on the UCI Adult Dataset to predict individuals' yearly income in the US and achieved a very high prediction accuracy of 88.16%.

# Results

This study reinforces the predictive power of machine learning models observed in prior economic research on income prediction. Compared to previous research in this area, this study makes several distinctive contributions. Firstly, it leverages both individual longitudinal (intra-individual) characteristics and population (inter-individual) characteristics in national longitudinal survey data, yielding enhanced accuracy and ROC area in predicting individual income levels. Secondly, asides from achieving satisfactory accuracy at predicting income levels, this study explores the significance levels of various socioeconomic factors in the prediction task, leading to valuable insights. With these unique aspects, this work sheds new light on the potential of employing longitudinal characteristics in research to enhance performance on individual income prediction, and providing valuable practical insights on the relationship between socioeconomic factors and individual income levels, for both individuals striving for financial success and policymakers seeking to promote broader societal well-being.

# Dataset

The dataset used in this study comes from the National Longitudinal Surveys with many cohorts tracking the life trajectories of several groups of American men and women. The NLS is a long-withstanding resource used by many economists and sociologists and should provide credible and reliable data for this research.

## NLSY97 Data

The NLSY97 cohort is purposely chosen for this investigation. "The NLSY97 consists of a nationally representative sample of 8,984 men and women born during the years 1980 through 1984 and living in the United States at the time of the initial survey in 1997." "The NLSY97 collects extensive information on respondents' labor market behavior and educational experiences. The survey also includes data on the youths' family and community backgrounds to help researchers assess the impact of schooling and other environmental factors on these labor market entrants." Following the initial survey, the cohort includes follow-up interviews and continuations, which is what this study will be looking at.

NLSY97 has a total of 87227 variables as of this study date, from which a small subset of variables on education, employment, demographic information, socioeconomic status are purposely chosen for this research, some of which having repeated measures across selected years. Table 1 lists the variables considered, where the income variable is the dependent variable to be predicted and all other variables are independent variables used to predict the dependent variable income. In this paper, independent variables and features are used interchangeably.

To utilize the temporal information within the longitudinal data, we include repeated measures of individuals from several different time points, spanning the years 2021, 2019, 2017, and 2015, consisting of a total of $8984 \times 4 = 35936$ entries.

Table 1. Variables Chosen from NLSY97 for This Study (all of numeric type)

| Variables | NLSY97 variable names | Measures in years |
|---|---|---|
| sex | KEY!SEX, RS SEX (SYMBOL) | 1997 |
| race | KEY!RACE, RACE OF R (SYMBOL) | 1997 |
| degree | HIGHEST DEGREE RECEIVED | XRND (Cross Round) |
| biological father's highest grade | BIOLOGICAL FATHERS HIGHEST GRADE COMPLETED | 1997 |
| biological mothers highest grade | BIOLOGICAL MOTHERS HIGHEST GRADE COMPLETED | 1997 |
| residential fathers highest grade | RESIDENTIAL FATHERS HIGHEST GRADE COMPLETED | 1997 |
| residential mothers highest grade | RESIDENTIAL MOTHERS HIGHEST GRADE COMPLETED | 1997 |
| parental household income | GROSS HH INCOME IN PAST YEAR | 2003 |
| highest grade | RS HIGHEST GRADE COMPLETED | 2021 |
| age | RS AGE IN MONTHS AS OF INTERVIEW DATE | 2015, 2017, 2019, 2021 |
| industry | YEMP, TYPE OF BUS OR INDUSTRY CODE (2002 CENSUS 4-DIGIT) 01 (ROS ITEM) | 2015, 2017, 2019, 2021 |
| occupation | YEMP, OCCUPATION/JOB CODE (2002 CENSUS 4-DIGIT) 01 (ROS ITEM) | 2015, 2017, 2019, 2021 |
| total work weeks | # WEEKS EVER WORKED AT JOB 01, 02, 03, 04, 05, 06, 07, 08 | 2015, 2017, 2019, 2021 |
| yearly work hours | YYYY EMPLOYMENT: TOTAL HOURS WORKED WEEK WW (YYYY is year and WW ranges from 01 to 52) | 2015, 2017, 2019, 2021 |
| income (yearly) | TOTAL INCOME FROM WAGES AND SALARY IN PAST YEAR | 2015, 2017, 2019, 2021 |

# Data Preprocessing

Data preprocessing plays a crucial role in preparing data for machine learning prediction tasks. The goal is to ensure the input data is structured appropriately for machine learning. In this study, data preprocessing serves several purposes.

First, it extracts longitudinal information from the original dataset, which is essential for understanding trends and patterns over time for longitudinal analysis. This process ensures that the temporal aspect of the data or the trajectories of individuals over time is preserved and can be effectively utilized in subsequent prediction tasks.

Second, it filters out invalid entries from the dataset, thus enhancing the quality and reliability of the data used in prediction tasks.

Lastly, it encodes the original data types into more suitable types when necessary to improve prediction accuracy. This includes, for example, converting categorical variables into nominal variables with one-hot encoding and scaling numerical variables to a common range, and properly labeling missing values.

## Extract and Represent Longitudinal Information

Longitudinal data and analysis have been extensively explored in classic statistical models (14) (15). In this study, we will explore the utilization of longitudinal data information in machine learning model settings.

To utilize the trajectory information of individuals over time in the longitudinal data, repeated measures of the same individual across different years are compiled into separate entries. In other words, the longitudinal information in the dataset is extracted out by encoding one entry for each individual with measures from N years into N entries for this individual, each for one year's measure of the individual, as shown in Table 2 (original entry with repeated measures over years) and 3 (encoded entries with one for each year). For new data in Table 3, each entry represents a unique observation, e.g., one individual at a specific time point.

With this data transformation, the time-varying information is captured in the by-individual by-time data structure, making it robust against missing values and irregularly spaced measures.

Table 2. Original Dataset Entry with Repeated Measures Over Years

| Variables | Entry 1 |
| --- | --- |
| individual id | 2 |
| sex | 1 |
| race | 5 |
| degree | 2 |
| biological father's highest grade | 17 |
| biological mothers highest grade | 15 |
| residential fathers highest grade | 14 |
| residential mothers highest grade | 15 |
| parental household income | 75000 |
| highest grade | 14 |
| age 2019 | 448 |
| industry 2019 | 7680 |
| occupation 2019 | 3820 |
| total work weeks 2019 | 646 |
| yearly work hours 2019 | 2700 |
| income (yearly) 2019 | 128400 |
| age 2021 | 472 |
| industry 2021 | 7680 |
| occupation 2021 | 3820 |
| total work weeks 2021 | 749 |
| yearly work hours 2021 | 3220 |
| income (yearly) 2021 | 115000 |

Table 3. Encoded Entries with One for Each Year

| Variables | Entry 1 | Entry 2 |
|---|---|---|
| individual id | 2 | 2 |
| sex | 1 | 1 |
| race | 5 | 5 |
| degree | 2 | 2 |
| biological father's highest grade | 17 | 17 |

| biological mothers highest grade | 15 | 15 |
| residential fathers highest grade | 14 | 14 |
| residential mothers highest grade | 15 | 15 |
| parental household income | 75000 | 75000 |
| highest grade | 14 | 14 |
| age | 448 | 472 |
| industry | 7680 | 7680 |
| occupation | 3820 | 3820 |
| total work weeks | 646 | 749 |
| yearly work hours | 2700 | 3220 |
| income (yearly) | 128400 | 115000 |

## Filter Out Invalid Entries

Out of 35936 entries, 15245 of them don't have valid income values (with negative values from NLSY97 data), which are removed in this study, leaving a total of 20691 entries. In order not to lose too many entries, we decide to keep the entries with missing values on independent variables (or non-income variables) and replace all missing values for nominal and numeric attributes in a dataset with the modes and means from the training data.

## Data Type Conversion and Feature Engineering

Table 1 lists the variables of NLSY97 raw data used in this study, all being numerical types with negative values -1 through -5 representing various not-available cases. Entries with invalid income values are filtered out in the previous step, while all negative values for other independent variables are marked as missing values for machine learning prediction tasks.
The following data type conversion and feature engineering are performed on the raw data, to improve the performance of machine learning models.

The dependent variable income is categorized into three nominal values throughout this study as shown in Table 4, making the prediction tasks in this study multi-class classification tasks.

Table 4. Data type Conversion and Feature Engineering

| Variables | Raw data type and values | New data type and values |
|---|---|---|
| sex | numeric: 1, 2 | nominal: 1, 2 |

| race | numeric: 1, 2, 3, 4, 5 | nominal: 1, 2, 3, 4, 5 |
|---|---|---|
| degree | numeric: 0, 1, 2, 3, 4, 5, 6, 7 | nominal: 0, 1, 2, 3, 4, 5, 6, 7 |
| biological father's highest grade | numeric: 1-20, 95 (ungraded) | numeric: 0-20, 95 |
| biological mothers highest grade | numeric: 1-20, 95 (ungraded) | numeric: 0-20, 95 |
| residential fathers highest grade | numeric: 1-20, 95 (ungraded) | numeric: 0-20, 95 |
| residential mothers highest grade | numeric: 1-20, 95 (ungraded) | numeric: 0-20, 95 |
| parental household income | numeric | numeric |
| highest grade | numeric: 1-20, 95 (ungraded) | numeric: 0-20, 95 |
| age | numeric | numeric |
| industry | numeric: 170-9990 | nominal: 1-18 |
| occupation | numeric: 10-9990 | nominal: 1-33 |
| total work weeks | numeric | numeric |
| yearly work hours | numeric | numeric |
| income (yearly) | numeric: 0-380288 | nominal:<br>- 1 (Less than 50K),<br>- 2 (between 50K and 100K),<br>- 3 (More than 100K) |

# Exploratory Analysis

After the data processing on the raw NLSY97 data, the new dataset contains 20691 entries and 15 numeric and nominal variables, 9 being numeric and 6 being nominal.

The goal of exploratory data analysis is to identify potential patterns and variables that can be useful in prediction tasks. For this purpose, a covariate analysis is conducted to examine the correlation between each independent variable and the dependent variable, income. The covariate analysis is performed on the survey data of 20691 entries. Subsequently, multivariate analysis to examine the collective influence of independent variables on the dependent variable is conducted with machine learning algorithms, as described in Methods section.

We use the Spearman correlation coefficients between all pairs of variables to identify independent variables with high correlation with the dependent variable. Then, we narrow down the set of independent variables by removing repeated highly correlated independent variables. We calculate the Spearman correlation coefficient for a pair of variables with Excel, first ranking the entries with RANK.AVG function, then calculating correlation coefficient with CORREL function on the ranking numbers.

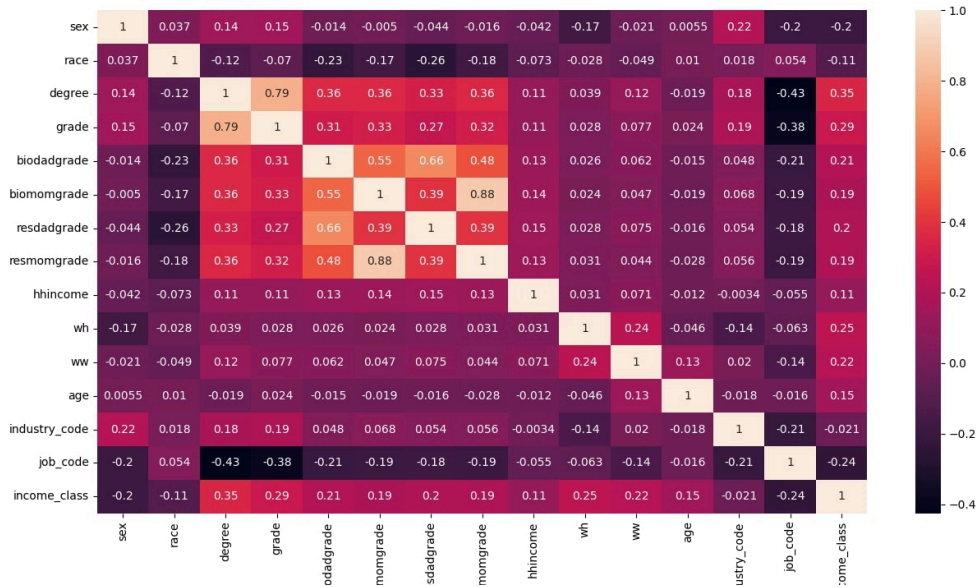

Figure 1. Heatmap of Spearman correlation coefficients for all variable pairs listed in Table 1

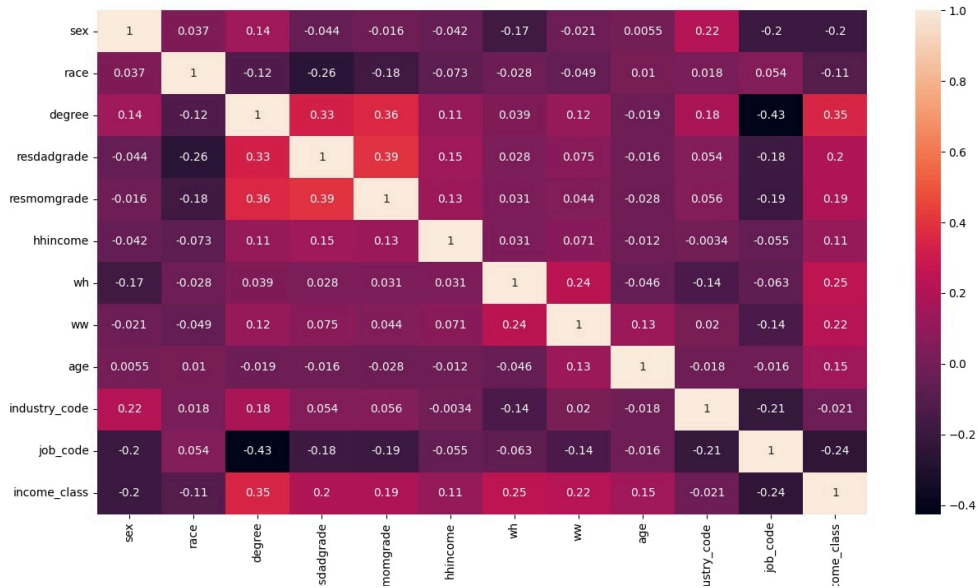

Figure 2. Updated heatmap of Spearman correlation coefficients for remaining 12 variables

Figure 1 displays the heatmap of Spearman correlation coefficients for all variable pairs listed in Table 1. There is high correlation between pairs (highest degree, highest grade), (biological father's highest grade, residential fathers highest grade) and (biological mothers highest grade, residential mothers highest grade), so we decide to keep one variable from each pair, that is, highest degree, residential fathers highest grade, and residential mothers highest grade, leaving us 12 variables for subsequent study as shown in Table 5. Figure 2 displays the updated heatmap of Spearman correlation coefficients for the remaining 12 variables.

Table 5. 12 Variables and Their Data Type Used in This Study

| Variables | Data type and values |
|---|---|
| sex | nominal: 1, 2 |
| race | nominal: 1, 2, 3, 4, 5 |
| degree | nominal: 0, 1, 2, 3, 4, 5, 6, 7 |
| residential fathers highest grade | numeric: 0-20, 95 |
| residential mothers highest grade | numeric: 0-20, 95 |
| parental household income | numeric |
| age | numeric |
| industry | nominal: 1-18 |
| occupation | nominal: 1-33 |
| total work weeks | numeric |
| yearly work hours | numeric |
| income (yearly) | nominal:<br>- 1 (Less than 50K),<br>- 2 (Between 50K and 100K),<br>- 3 (More than 100K) |

# Methods

In this section, the predictive task is to classify provided input into predefined income categories. Compared to the covariate analysis conducted in the exploratory analysis, a multivariate analysis is performed to fully explore the collective influence of all independent variables on the dependent variable.

In this study, Weka, an open-source machine learning software toolkit developed by the University of Waikato in New Zealand, is utilized. Weka offers a diverse array of machine learning algorithms, encompassing classification, regression, and more.

# Models

A baseline model is first established to set a benchmark for further analysis. Then, the effectiveness of several popular machine learning algorithms in weka is assessed for income classification. This includes the multilayer perceptron neural network (16), sequential minimal optimization (17) for support vector machines (18) and random forest (19) - three powerful and representative machine learning models. Each model has proven effective in conducting prediction tasks in prior research. The top-performing algorithm is then selected to conduct additional prediction tasks, to facilitate further in-depth analysis.

## Baselining the Data

In exploratory analysis, the income classes show a distribution of 57.564%, 31.344% and 11.092% respectively for the three classes (<50K, >=50K and <100K, and >=100K). The baseline model is to predict the majority class <50K for all items, which gives us the baseline in Table 6.

Table 6. Benchmark by Baseline Model

| Models | Correctly Classified Instances | ROC Area |
|---|---|---|
| vote majority | 57.6% | 0.500 |

## Multilayer Perceptron Neural Network

A multilayer perceptron (MLP) is a type of artificial neural network consisting of multiple layers of neurons. The neurons in the MLP typically use nonlinear activation functions, allowing the network to learn nonlinear relationships in data.

## Sequential Minimal Optimization for Support Vector Machine

Sequential Minimal Optimization (SMO) is a popular algorithm used for training Support Vector Machines (SVMs). SVMs are powerful supervised learning models for solving classification and regression tasks by finding the hyperplane that best separates the data points into predefined classes while maximizing the margin between the classes. SMO, developed by John Platt, can efficiently solve the quadratic programming problem for training SVMs.

## Random Forest

Random Forest consists of a group of decision trees, each being trained independently on a random subset of the training data and variables. For predictions, the final result is the average (for regression tasks) or majority vote (for classification tasks) over the independent predictions from all the trees in the forest.

Random Forest has many advantages in prediction tasks, including high predictive accuracy, resistance to overfitting, robustness to noise and outliers, adeptness at handling missing values, and scale well.

# Analysis

## Model analysis

For fair and consistent model performance analysis, we adopt a uniform test option - percentage split - in the classification tasks in this study. This involves allocating eighty percent of the data for model training and the remaining twenty percent for model testing. This train-test split helps detect and prevent model overfitting. We also conducted the analysis using 10-fold cross-validation and got very similar results.

The evaluation metrics used to assess model performance include prediction accuracy in Equation (1) (measured by Correctly Classified Instances metrics in weka) and Area Under the Curve (AUC - a weighted average of ROC Area in weka) for classification tasks.

$$\begin{aligned}\text{accuracy} &= \frac{\text{number of correctly classified instances}}{\text{total number of instances}} \times 100\% \\ &= \frac{\text{number of true positive instances} + \text{number of true negative instances}}{\text{total number of instances}} \times 100\%\end{aligned} \quad (1)$$

The Receiver Operating Characteristic (ROC) curve is a graphic representation of the performance of a classification model across different threshold values. The AUC measures the area under the ROC curve. A perfect classifier would have an AUC of 1 and a random classifier would have an AUC of 0.5, which is no better than chance.

## Longitudinal Data Analysis

To evaluate the influence of longitudinal data on prediction performance, two classification tasks are set up: one using NLSY97 data solely from the year 2021 and the other using NLSY97 data from multiple years including 2021, 2019, 2017 and 2015. Then we assess the model performance metrics for both tasks to determine whether integrating longitudinal data leads to potential improvements in the model prediction performance.

## Factor Significance Analysis

Compared to traditional statistical methods, machine learning models usually lack a clear interpretation of significance levels of individual factors in prediction tasks. SHapley Additive exPlanations (SHAP) values (20) are used for factor significance analysis in machine learning models in this study. SHAP is a game theoretic approach to explain the output of machine learning models. It assigns each feature an important value - a SHAP value. SHAP values quantify the contribution of individual features to model predictions, enhancing the interpretability of complex models and helping identify the most influential features in the model's decision-making process. In this study, the Tree SHAP algorithm is used in calculating SHAP values, which is tailored for tree-based algorithms like random forests and doesn't rely on feature-independence assumption.

# Results and Discussion

In this section, we compile the analysis results for model selection, longitudinal data contribution and factor significance. Weka is used in classification tasks with different machine learning models.

## Model Selection

To assess the effectiveness of the three machine learning algorithms for the classification tasks in this study, NLSY97 data from the year 2021, 2019, 2017 and 2015 is fed into each of them. Table 7 gives the prediction accuracy and ROC area for the three algorithms in this prediction task. Random forest stands out with both higher prediction accuracy and larger ROC area, which is chosen for all the subsequent analysis.

Table 7. Model Performance Comparison

| Models | Correctly Classified Instances | ROC Area |
| --- | --- | --- |
| MLP | 68.0 % | 0.807 |
| SVM | 67.9 % | 0.732 |
| Random Forest | 72.6 % | 0.849 |

For the testing dataset consisting of twenty percent of the input data (20% of 20691, i.e., approximately 4138 instances), the three machine learning models produce the confusion matrix presented in Table 8 through Table 10.

Table 8. Confusion Matrix for MLP Model

| Actual class ↓  Predicted class → | Less than 50K | Between 50K and 100K | More than 100K |
| --- | --- | --- | --- |
| Less than 50K | 1888 | 460 | 34 |
| Between 50K and 100K | 435 | 762 | 100 |
| More than 100K | 74 | 222 | 163 |

Table 9. Confusion Matrix for SVM Model

| Actual class ↓  Predicted class → | Less than 50K | Between 50K and 100K | More than 100K |
| --- | --- | --- | --- |
| Less than 50K | 2036 | 338 | 8 |
| Between 50K and 100K | 538 | 739 | 20 |
| More than 100K | 81 | 342 | 36 |

Table 10. Confusion Matrix for Random Forest Model

| Actual class ↓   Predicted class → | Less than 50K | Between 50K and 100K | More than 100K |
|---|---|---|---|
| Less than 50K | 2064 | 299 | 19 |
| Between 50K and 100K | 473 | 749 | 75 |
| More than 100K | 87 | 180 | 192 |

## Longitudinal Data Contribution

The following experiments are designed to examine the contribution of longitudinal data to the model performance in predicting individual income.

Five random forest models are trained as listed in Table 11. Four models are trained with data from individual years, while the fifth model is trained with data points from multiple years to utilize longitudinal data.

Table 11. Five Models Trained

| Models | Training Data |
|---|---|
| Model 2015 | 4000 random data points from year 2015 |
| Model 2017 | 4000 random data points from year 2017 |
| Model 2019 | 4000 random data points from year 2019 |
| Model 2021 | 4000 random data points from year 2021 |
| Model 2015-2021 | 4000 random data points from year 2015-2021, 1000 data points from each year |

Four sets of test data from the year 2015 through 2021 respectively are used to evaluate the model performance. Each set consists of 1000 test data points randomly drawn from each year excluding those used in model training. Table 12 gives the prediction accuracy and ROC area of the random forest machine learning models with and without longitudinal data on the four sets of test data. The fifth Model 2015-2021 trained with longitudinal data outperformed other models trained without longitudinal data, in terms of both correctly classified instances and ROC areas.

Table 12. Comparison Prediction Performance with and without Longitudinal Data

| Tasks | Model Performance without Longitudinal Data | Model Performance with Longitudinal Data |
|---|---|---|

| 1000 test data points from 2015 | Model: Model 2015<br>Correctly Classified Instances: 70.3%<br>ROC Area: 0.768 | Model: Model 2015-2021<br>Correctly Classified Instances: 72.2%<br>ROC Area: 0.801 |
|---|---|---|
| 1000 test data points from 2017 | Model: Model 2017<br>Correctly Classified Instances: 65.1%<br>ROC Area: 0.772 | Model: Model 2015-2021<br>Correctly Classified Instances: 67.6%<br>ROC Area: 0.810 |
| 1000 test data points from 2019 | Model: Model 2019<br>Correctly Classified Instances: 65.3%<br>ROC Area: 0.795 | Model: Model 2015-2021<br>Correctly Classified Instances: 68.4%<br>ROC Area: 0.819 |
| 1000 test data points from 2021 | Model: Model 2021<br>Correctly Classified Instances: 64%<br>ROC Area: 0.793 | Model: Model 2015-2021<br>Correctly Classified Instances: 65.9%<br>ROC Area: 0.813 |

The result demonstrates a significant improvement in model prediction performance with the use of longitudinal data, confirming our initial expectation that incorporating longitudinal data can enhance machine learning models for individual income prediction.

## Factor Significance

Figure 3 and 4 show the random forest feature importance with SHAP values. In these graphs, we can see how each factor impacts the model output respectively.

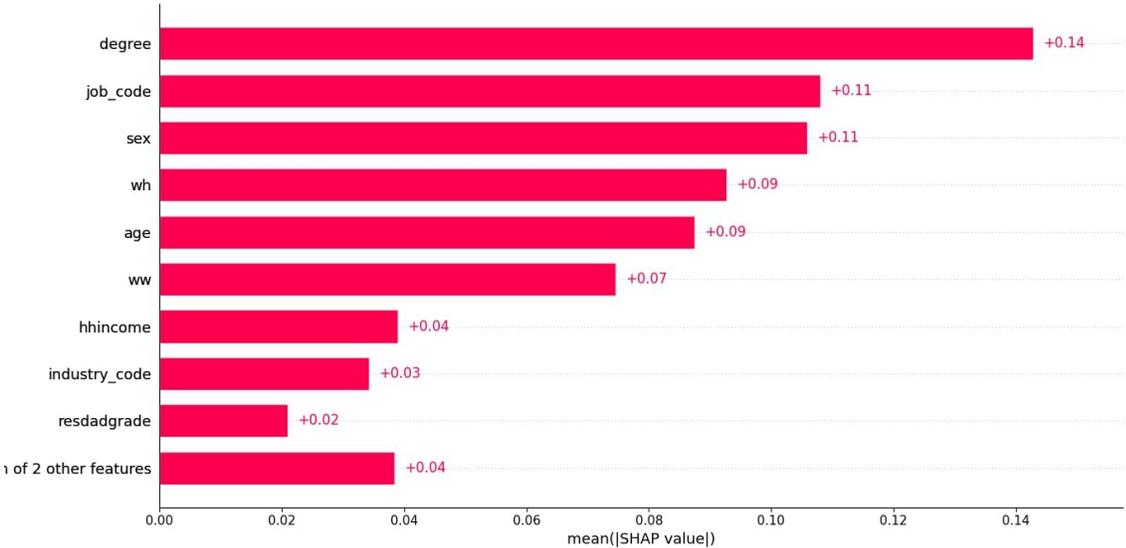

Figure 3. SHAP value (impact on model output) - mean absolute

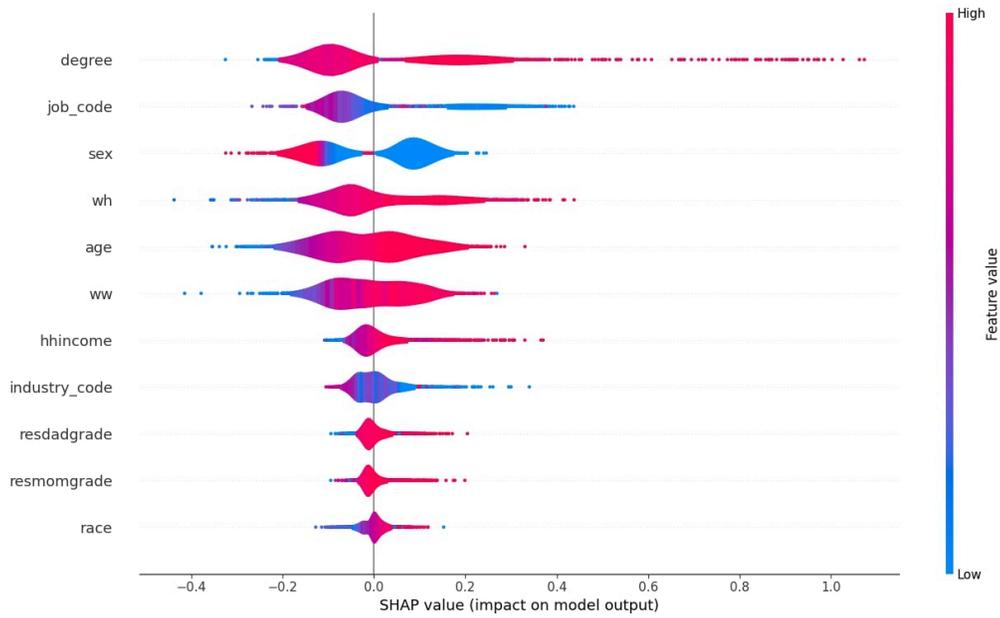

Figure 4. SHAP value (impact on model output)

In Figure 3, the most influential variables for individual income are an individual's highest education degree, occupation and sex, in that order. Following closely are the variables of yearly working hours, age and work tenure, with all other variables considered less influential.

In Figure 4, it's not surprising to observe that variables such as the highest degree attained, yearly working hours, age, work tenure, and parental household income exhibit a positive correlation with individual income. Additionally, specific occupations are associated with higher incomes, and males tend to earn higher incomes compared to females.

In addition to examining SHAP values for individual features, we conduct the following experiments to directly assess the quantitative impact of each individual feature on income prediction performance. We establish a baseline using all 11 features, then systematically drop one feature at a time for 11 additional experiments. The prediction performance of the random forest machine learning model for all 12 experiments is presented in Table 13, which aligns with the SHAP evaluation results pretty well.

Table 13. Prediction Performance with Different Feature Sets

| Experiments | Correctly Classified Instances | ROC Area |
| --- | --- | --- |
| baseline: with all 11 features | 72.6 % | 0.849 |
| without degree | 69.5 % | 0.825 |
| without occupation | 70.1 % | 0.818 |
| without sex | 71.6 % | 0.840 |
| without yearly working hours | 71.8 % | 0.844 |
| without age | 71.4 % | 0.844 |
| without work tenure | 72.2 % | 0.841 |
| without parental household income | 71.3 % | 0.839 |
| without industry code | 70.2 % | 0.831 |
| without residential father's grade | 72.3 % | 0.844 |
| without residential mother's grade | 71.9 % | 0.843 |
| without race | 71.0 % | 0.839 |

# Conclusion and Future Work

This study demonstrates that with careful data preprocessing, feature engineering, proper model selection, and effectively leveraging longitudinal data, machine learning methodologies can predict individual income with satisfactory performance. It also provides valuable insights on how various socioeconomic factors influence individuals' income levels, aiding individuals striving for financial success in maintaining focus and assisting policymakers in making informed decisions for the broader socioeconomic society.

This study confirms the effectiveness of machine learning models in predicting individual income. It also has several unique findings. Firstly, making proper use of longitudinal information in input dataset yields enhanced prediction performance in individual income prediction tasks. Secondly, the analysis makes use of SHAP values and auxiliary approaches, leading to valuable insights into the significance levels of various socioeconomic factors on individual incomes, with potentially significant impact in practice.

This study identifies the top three influential variables for individual incomes as an individual's highest education degree, occupation and sex, followed by yearly working hours, age and work tenure. The actionable variables, such as highest degree, occupation, and yearly working hours, hold particular significance for individuals aiming for higher incomes.

There are some interesting future directions for this research.

- This study focuses on the NLSY97 data, while similar methodologies employed herein can be extended to other datasets.
- Additional socioeconomic factors, beyond those listed in Table 1, such as health-related and geographic factors, could be included to discover more patterns and influential factors affecting individual incomes.

- When removing the sex/gender variable from the machine learning model (as listed in Table 5), lower prediction accuracy (drops to 71.5805% from 72.6196 %) and smaller ROC area (drops to 0.84 from 0.849) are observed. These results highlight the significant issue of sex/gender bias within data and machine learning processes. This bias can manifest in various ways, including feature selection, algorithmic biases during modeling and others. Future exploration in this direction has the potential to discover interesting findings and promote fairness, equity, and transparency in machine learning practices.
- This study exploits longitudinal information in the input dataset by disaggregating repeated measures on an individual into multiple entries, each corresponding to a specific time point for the individual, before feeding into machine learning models. There may be alternative methods for leveraging longitudinal data.
- One limitation with the use of longitudinal data in income prediction is that due to the inflation of individual incomes over years, employing longitudinal data over an extensive time frame might lead to a deterioration in the accuracy of prediction tasks. There may be methods to mitigate the impact of income inflation in input data, allowing for more effective use of longitudinal data.

# Acknowledgements

I would like to extend my heartfelt thanks to Professor Ramin Ramezani from UCLA for the informative and engaging meetings on AI and Machine Learning, which have further nurtured my interest in the field. I also want to express my gratitude to my TA, Joanna Gilberti from New York University, for her invaluable assistance in editing and maintaining progress with this project.